\documentclass[10pt]{article}
\usepackage[preprint]{tmlr}
\usepackage{amsmath,amssymb}
\usepackage{booktabs,array,longtable}
\usepackage{graphicx}
\usepackage{microtype}
\usepackage{placeins}
\usepackage{xcolor}
\usepackage{hyperref}
\usepackage{url}
\hypersetup{colorlinks=true,linkcolor=blue!45!black,citecolor=blue!45!black,urlcolor=blue!55!black}
\graphicspath{{figures/}}

\title{Local Gains and Fixed-Assignment Set Losses in Shared Set Decoders}

\author{
\name Ze Zhang$^{1}$, Yang Zhang$^{2}$\\
\addr $^{1}$ School of Biomedical Engineering, ShanghaiTech University, Shanghai, China\\
\addr $^{2}$ Wuhan United Imaging Surgical Co., Ltd., Wuhan, China
}
\def\month{08}
\def\year{2026}
\def\openreview{}

\begin{document}
\maketitle

\begin{abstract}
A query-relation deletion can improve the edited slot while reducing the utility
of the prediction set that contains it. We study this tension in two related
ResNet-50 DETR-family checkpoints using recorded, selection-conditional evidence
from 710 paired image--relation units per checkpoint. The primary comparison
subtracts a matched active control, which deletes the same leader source at a
different recorded recipient, from the selected target deletion. It is
therefore a composite contrast rather than a same-recipient placebo.

The target-minus-control contrast is locally positive and fixed-assignment
negative in both checkpoints. The opposite-sign pattern occurs within 302/710
DETR units and 460/710 DINO units. After rematching, the corresponding counts
are 285/710 and 433/710. Rematching and native selection absorb enough of the
mean loss for DETR intervals to cross zero, whereas DINO intervals remain
negative, so persistence across readouts differs by checkpoint. A fixed-map
comparison between hard deletion and a mass-preserving edit also differs before
rematching. That comparison is conditional on the outcome-blind map and does
not establish same-dose transport.

Local intervention success therefore does not determine the consequence for a
jointly decoded set. The supported conclusion is selection-conditional deletion
sensitivity whose persistence depends on the readout and intervention operator.
We do not identify an intervention-invariant edge mechanism, detector-level
degradation, population prevalence, or the value of a training-time regularizer.
\end{abstract}

\paragraph{Keywords.}
Set prediction; object queries; mechanistic interpretability; attention
intervention; matching-aware readout; intervention sensitivity.

\section{Introduction}
\label{sec:introduction}

An edit to one part of a model can improve the output nearest to the edit while
making the larger prediction worse. This possibility is especially
consequential in a shared set decoder: object queries are updated jointly, and
the final output is determined through assignment and selection rather than by
reading each slot in isolation \citep{carion2020detr,zhang2023dino}. A positive
local response can therefore coexist with a loss in the utility of the fixed
prediction set. The scientific problem is not merely whether the edit changes
the model, but how that change propagates across the set and which readout
captures it.

Intervention studies often use deletion, corruption, or patching to reason about
a component's role. Prior work has shown that such conclusions depend on both
the intervention and the metric used to evaluate it
\citep{jain2019attention,wiegreffe2019attention,zhang2024patching,makelov2024subspace}.
Set prediction adds a further complication. The same edited computation can be
evaluated at the target slot, across other slots under the original assignment,
after Hungarian rematching, or after native output selection. These stages are
not interchangeable: reassignment can recover utility without undoing the
underlying state change, and native selection can expose another response.

We ask: \emph{what does a reproducible query-relation deletion response
establish when its comparator, selected population, intervention operator, and
readout stage are made explicit?} This question separates an observed response
from stronger interpretations. Hard deletion, attenuation, and mass-preserving
replacement are different versions of an intervention unless a mapping between
them is established \citep{vanderweele2013versions,massidda2023soft}. Likewise,
a fixed-assignment loss is not detector average precision, and a response within
a selected population is not a prevalence estimate over arbitrary relations.

\begin{figure*}[t]
\centering
\includegraphics[width=0.95\linewidth]{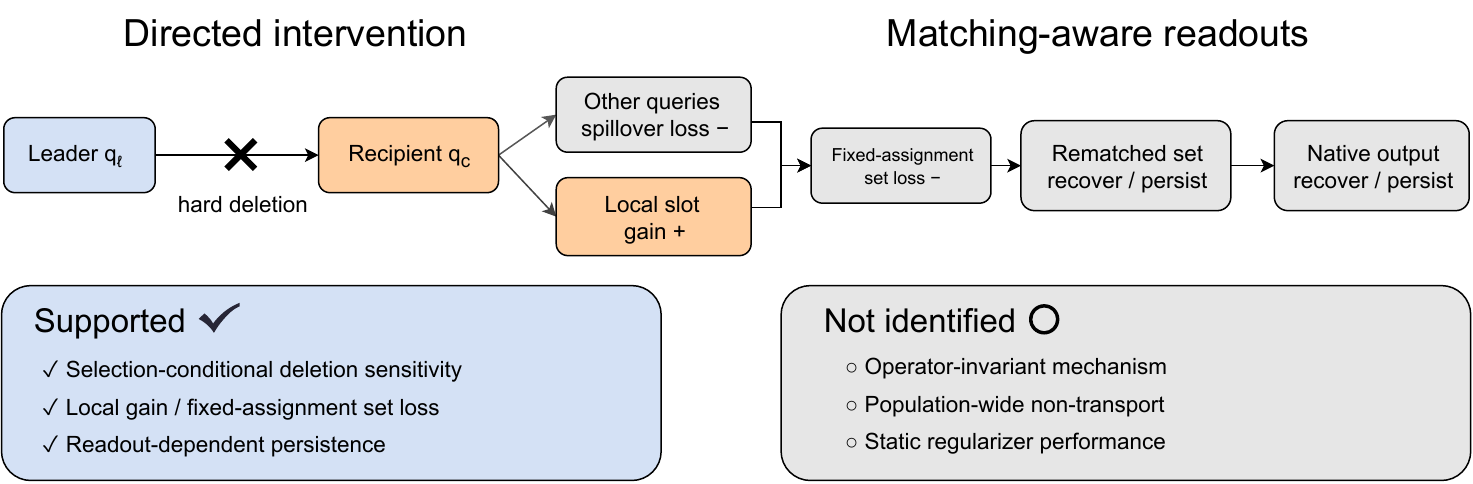}
\caption{\textbf{A local query gain can coexist with a fixed-assignment set
loss.} A
selected directed relation is hard-deleted in a jointly decoded query set and
compared with a different-recipient active control. The target-slot response can
be positive while fixed-assignment set utility is negative; fixed, rematched,
and native readouts then show how much of that response persists. The lower
panels separate the supported selection-conditional findings from mechanism,
population-transport, and regularizer claims that the evidence does not
identify.}
\label{fig:concept}
\end{figure*}

Our core insight is that the local effect, the set-level effect, and the
persistence of that effect after matching are linked observations of one
jointly decoded response. Collapsing them into a single verdict about whether a
relation is helpful or harmful would discard that structure. We implement this
view with a fixed target-minus-matched-active-control comparison, a progression
from local to fixed, rematched, and native readouts, and a separate fixed-map
comparison of hard deletion with a mass-preserving edit. Supporting dose,
fresh-context, and design-authorization analyses then test whether the evidence
permits a stronger interpretation. Figure~\ref{fig:concept} summarizes this
progression and its claim boundary.

The primary contrast agrees across the two checkpoints, but its persistence
does not. In both checkpoints, target deletion relative to the active control
improves the local target slot while reducing fixed-assignment set utility, and
opposite-sign responses occur in many selected units. After rematching and
native selection, the DETR intervals cross zero while the DINO intervals remain
negative. The mapped hard-minus-mass difference is already visible before
rematching, yet the available evidence does not calibrate operator transport or
select a unique mechanism.

This paper makes three contributions. First, it establishes an empirical local--
set sign reversal for the selected deletion contrast in two evaluated checkpoints,
at aggregate and within-unit levels. Second, it characterizes how the response
persists through fixed assignment, rematching, and native selection, exposing
checkpoint-specific recovery. Third, it develops an intervention-specific
interpretation of these findings: the mapped operator comparison and stopping
analyses show why the observed sensitivity does not itself authorize transport,
mechanism, or training-design claims.

\section{Related Work}
\label{sec:related}

\subsection{Intervention responses and mechanism claims}

Attention-faithfulness and activation-intervention studies have questioned when
a successful corruption or patch identifies the semantic role of a model
component. The relevant lesson here is that the intervention and evaluation
metric are part of the measured claim, rather than neutral implementation
details
\citep{jain2019attention,wiegreffe2019attention,zhang2024patching,makelov2024subspace,
guo2026igsd,quirke2026transfer,joshi2026generalise}. We apply that discipline to
a jointly decoded set, where a local readout and a set-level readout can disagree.

\subsection{Query interaction as a decoder design problem}

Work on object detection has treated query interaction, denoising, grouping,
relation bias, routing, and matching stability as architecture or training
problems
\citep{li2022dndetr,chen2023groupdetr,gao2024ease,senthivel2024qrdetr,
hou2024relation,zhang2026dualr,liu2023stable}. These studies motivate the
importance of interaction among queries. Our study addresses a different
question at fixed checkpoints: how a selected directed-relation intervention
appears across assignment-sensitive readouts, and what that response licenses
scientifically.

\subsection{Matching-aware readout}

Set prediction makes assignment part of measurement. Evaluating the edited
model under the original assignment preserves one correspondence; rematching
permits the set to reorganize; native selection evaluates the model's final
output rule. We use these readouts to locate where an intervention response
persists. The contribution is not a new matching procedure, but an empirical
account that keeps the stages separate while comparing a selected deletion with
an active control and an indexed alternative operator.

\section{Study Design and Estimands}
\label{sec:formulation}

\subsection{Datasets, checkpoints, and evaluated populations}

The primary discovery population comes from a validation-derived Open Images
partition with a fixed COCO-category mapping. We evaluate two fixed
ResNet-50 checkpoints separately: DETR-R50-500 and DINO-R50 4-scale 12-epoch.
Their checkpoint SHA-256 values are
\texttt{e632da11...784eadc} and \texttt{0bcd6b0c...307955a}, respectively.
The same discovery image identities are used for both checkpoints, but the
selected query relation is model-specific. Agreement between the checkpoints
is therefore controlled checkpoint replication, not architecture-family
replication.

The statistical unit is one image with one selected directed relation. The
discovery partition contains 1,500 images. Requiring the target pair and all
declared controls to be feasible in both checkpoints reduces 752 jointly
pair-eligible images to 710 complete paired units per checkpoint. These units
are not random query edges: every estimate is conditional on the baseline-only
selection procedure below. A separate supporting assay uses COCO 2017
validation. It has 247 eligible DETR images and 252 eligible DINO images and
uses the first 128 complete units per checkpoint in its recorded stratified
order. The primary quantitative claims use only the 710-unit discovery
populations.

\subsection{Relation selection and matched active control}

For a fixed primary ground-truth object \(g\), checkpoint \(m\), and query
\(q\), query quality is
\begin{equation}
r_m(q,g)=p_m(c_g\mid q)\operatorname{IoU}(b_q,b_g),
\label{eq:quality}
\end{equation}
where \(p_m\) is the DETR softmax class probability after excluding the
no-object class or the DINO sigmoid class score. The discovery primary object
is the retained mapped object selected by the recorded seeded-minimum-hash
rule. For the COCO supporting assay, noncrowd objects are ordered by decreasing
pixel area and then annotation ID.

Queries with IoU below 0.30 are ineligible. Among the remaining queries, the
leader \(q_\ell\) and competitor \(q_c\) are the highest- and second-highest
quality queries, with ascending query ID breaking ties; a strictly positive
quality margin is required. An eligibility sham must be distinct from both,
have target IoU below 0.05, and minimize the absolute target-class-score
difference from the competitor, again with query ID as the tie break. If any
condition fails, the unit is rejected rather than admitted under a changed
threshold.

The discovery active control uses a separately matched recipient \(q_h\).
Among nonleader and noncompetitor queries with target IoU below 0.05, it
minimizes standardized Euclidean distance to the competitor over target-class
score, layer-3 hidden-state \(\ell_2\) norm, and leader-to-source attention.
Candidate-pool standard deviations scale the three features and query ID breaks
ties. The target arm deletes \(q_c\leftarrow q_\ell\); the active-control arm
deletes \(q_h\leftarrow q_\ell\). Both selections use baseline quantities and
do not read intervention outcomes. Because the recipients differ, the primary
target-minus-control estimand is a composite active comparison rather than a
same-recipient placebo or an exchangeability claim.

\subsection{Intervention operators}

Hard deletion acts on the decoder input layer 3 pre-softmax attention-logit
tensor in every head. It sets the selected logit
\(z_{h,c,\ell}\leftarrow-\infty\) before softmax. The remaining logits are
unchanged at the intervention point, while renormalization redistributes the
deleted attention mass over the remaining sources.

The mass-preserving operator acts after softmax. At the target recipient it
exchanges the selected target and donor weights while preserving their
two-cell sum; the corresponding control uses the matched recipient and the
competitor donor. It does not preserve value content or later decoder state.
The conditional operator analysis maps each hard-deletion unit to one of the
recorded mass conditions using two input-only quantities: absolute target-weight
change and immediate message norm. After a \(\log(1+x)\) transform, the two
quantities are standardized by their model-specific pooled standard deviations;
Euclidean distance chooses the closest condition, with the smaller dose
breaking ties. Outcomes are not used to construct this map. The reported
hard-minus-mass contrast is conditional on the fixed map and is not a
same-dose or transport estimate.

\subsection{Utility and matching-aware readouts}

For each prediction--ground-truth pair, quality is the class score times IoU
from Eq.~\ref{eq:quality}. A deterministic Hungarian assignment maximizes the
sum of this quality under an injective prediction-to-object constraint. When
there are fewer predictions than objects, zero-valued dummy predictions are
added; exact ties are broken by prediction and object indices. Image-set
utility is mean assigned quality over ground-truth objects, with unmatched
objects contributing zero.

Let \(U_0\) be active-control image-set utility under its assignment. Let
\(U_1\) replace only the focal prediction row with its target-arm row,
\(U_2\) use all target-arm rows under the active-control assignment,
\(U_3\) rematch all target-arm rows, and \(U_4\) apply native output selection
before rematching. The decomposition is
\begin{equation}
\mathrm{focal}=U_1-U_0,\quad
\mathrm{spillover}=U_2-U_1,\quad
\mathrm{matching}=U_3-U_2,\quad
\mathrm{selection}=U_4-U_3.
\label{eq:readouts}
\end{equation}
Fixed-assignment, rematched, and native set effects are \(U_2-U_0\),
\(U_3-U_0\), and \(U_4-U_0\). The local endpoint is the selected recipient
slot. The target-set endpoint is the maximum quality among queries assigned to
the same target object. These endpoints locate where the observed response
persists; they are not interchangeable outcomes or separate causal mechanisms.

\subsection{Statistical uncertainty and multiplicity}

The two checkpoints are analyzed separately. All bootstrap analyses use 10,000
deterministic, hash-seeded resamples of paired images within the recorded
strata. The primary common-estimand analysis uses within-model max-studentized
simultaneous 95\% intervals over its declared reaggregation family. The
conditional mapped analysis uses the same form of simultaneous interval over
its eight readouts, with the input-only map held fixed outside the bootstrap.
Mapping uncertainty is therefore not propagated.

Within-unit sign events require local response \(>10^{-4}\) and the named set
response \(<-10^{-4}\). Their proportions use two-sided Wilson 95\%
intervals. These event rates and all mean effects describe only the selected
populations. The separate COCO assay uses simultaneous intervals and
bands fixed before outcome inspection: \(0.005\) (0.5 percentage point) for
query-local quality and \(0.001\) (0.1 percentage point) for per-object set
utility. These are minimum scientifically meaningful decision bands, not
estimates of a null effect. An interval inside a band is not treated as
evidence about the discovery effect unless the assay also passes its
hard-deletion sensitivity prerequisite; because that prerequisite fails, the
band classifications do not support a mechanism claim.

\subsection{Reproducibility artifacts}

The anonymous companion package contains per-image analysis-ready rows,
population and pair manifests, intervention and statistical configurations,
integrity checks, expected aggregates, and a CPU-only reproduction script.
The script reconstructs the reported discovery, dose, conditional mapped,
and fresh-context aggregates without model inference. The package excludes image
pixels and model weights and therefore supports exact aggregate reconstruction,
not end-to-end inference replication. The original runtime used Python 3.12.13,
PyTorch 2.10.0+cu128, torchvision 0.25.0+cu128, NumPy 1.26.4, SciPy 1.17.1,
and an NVIDIA RTX 4090; aggregate reproduction requires only Python and NumPy.

\section{Results}
\label{sec:results}

\subsection{Target deletion improves the local slot but reduces fixed-assignment set utility}
\label{sec:primary}

\begin{figure*}[t]
\centering
\includegraphics[width=0.96\linewidth]{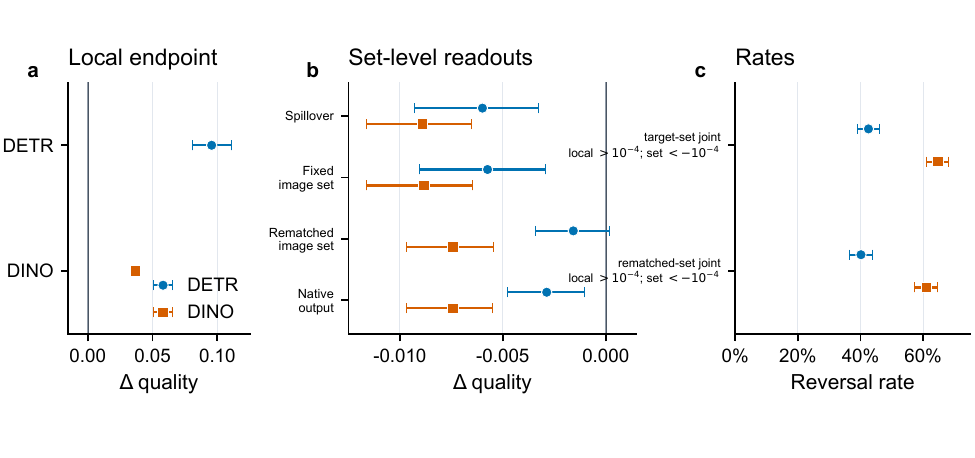}
\caption{\textbf{Primary selected-deletion result.} The target-minus-matched-
active-control contrast is locally positive and fixed-assignment negative in
both selected 710-unit populations, with frequent target-set and rematched
within-unit sign reversals. DETR and DINO are shown separately. The active
control is composite because target and control edits use different recipients;
the rates are selection-conditional and are not arbitrary-relation prevalence.}
\label{fig:hard}
\end{figure*}

The target-minus-matched-active-control contrast is local-positive and
fixed-assignment-negative in both checkpoints (Figure~\ref{fig:hard} and
Table~\ref{tab:primary}). For DETR, the local effect is $+0.095621$ with
simultaneous 95\% interval $[+0.070160,+0.121082]$, while the fixed-assignment
effect is $-0.005741$ with interval $[-0.010889,-0.000592]$. For DINO, the
corresponding local effect is $+0.036633$ with interval
$[+0.030623,+0.042643]$, and the fixed-assignment effect is $-0.008833$ with
interval $[-0.013130,-0.004536]$.

The opposite signs show that the recipient-slot response does not summarize the
consequence for the jointly decoded set. The intervention can improve the
selected slot while changes distributed across the fixed assignment produce a
negative image-level total. The primary evidence does not determine which
downstream interaction produces that total; it establishes that the local and
fixed-assignment readouts disagree under the implemented comparison. The sign
reversal is the empirical result. Its downstream mechanism remains unresolved.

On the selected populations and under the composite active control, the set
utility here is not detector average precision. The negative fixed-assignment
contrast therefore speaks to this set-level readout, not to detector-level
degradation or to whether suppressing the relation during training would
improve the model.

\begin{table*}[t]
\centering
\small
\caption{\textbf{Comparator-aligned primary results on the selected discovery
population.} Each checkpoint has 710 paired image--relation units. Target--control
is a composite contrast because target and control edits use different recipients.
Intervals are within-model simultaneous 95\% intervals; event rows use Wilson
intervals and the declared $10^{-4}$ deadzone. All primary quantitative claims
use the common estimand definitions specified in Section~\ref{sec:formulation}.}
\label{tab:primary}
\begin{tabular}{llrrr}
\toprule
Model & Endpoint & Estimate/count & Lower & Upper \\
\midrule
DETR & Target--control local & $+0.095621$ & $+0.070160$ & $+0.121082$ \\
DETR & Target--control fixed set & $-0.005741$ & $-0.010889$ & $-0.000592$ \\
DETR & Target--control rematched & $-0.001575$ & $-0.004544$ & $+0.001393$ \\
DETR & Target--control native & $-0.001686$ & $-0.004741$ & $+0.001370$ \\
DETR & Local $+$ / target-set $-$ & $302/710$ & $38.95\%$ & $46.20\%$ \\
DETR & Local $+$ / rematched $-$ & $285/710$ & $36.60\%$ & $43.79\%$ \\
DINO & Target--control local & $+0.036633$ & $+0.030623$ & $+0.042643$ \\
DINO & Target--control fixed set & $-0.008833$ & $-0.013130$ & $-0.004536$ \\
DINO & Target--control rematched & $-0.007424$ & $-0.010855$ & $-0.003992$ \\
DINO & Target--control native & $-0.007429$ & $-0.010861$ & $-0.003998$ \\
DINO & Local $+$ / target-set $-$ & $460/710$ & $61.20\%$ & $68.21\%$ \\
DINO & Local $+$ / rematched $-$ & $433/710$ & $57.35\%$ & $64.51\%$ \\
\bottomrule
\end{tabular}
\end{table*}

\subsection{Reassignment shows checkpoint-specific recovery}

The aggregate contradiction is also present within individual selected units.
Local-positive/target-set-negative events occur in 302/710 DETR units and
460/710 DINO units. When the negative condition is evaluated after rematching,
the counts are 285/710 and 433/710, respectively. These events use the declared
deadzone and show that the opposite-sign response is not produced only by
averaging positive and negative units.

Later readouts distinguish the checkpoints. DETR's target-minus-control
rematched and native means are $-0.001575$ and $-0.001686$, and both intervals
cross zero. DINO's corresponding means are $-0.007424$ and $-0.007429$, and both
intervals remain negative. Reassignment and native selection thus absorb enough
of DETR's mean fixed-assignment loss to make its later-readout intervals include
zero, while DINO's negative mean persists.

These rates describe the selected units only; they do not estimate occurrence
across arbitrary relations, detectors, or an architecture family. The narrower
finding is that the readout stage changes the observed persistence of the same
selected intervention response.

\subsection{A mapped operator difference appears before rematching}
\label{sec:operator}

The conditional hard-minus-mass comparison separates intervention versions
under the fixed outcome-blind map. In DETR, the mapped local, spillover, and
fixed-assignment means are $+0.081169$, $-0.005091$, and $-0.004981$.
Their simultaneous intervals exclude zero, whereas the rematched and native
intervals cross zero. In DINO, the corresponding local, spillover, fixed,
rematched, and native means are $+0.034817$, $-0.007747$, $-0.007701$,
$-0.006561$, and $-0.006571$; all five intervals exclude zero under the
within-model simultaneous family.

Because the contrast is visible in the local, spillover, and fixed-assignment
readouts, the mapped difference is not created solely by final rematching. The
data are consistent with the two operators altering normalization, attention
redistribution, donor content, or later shared computation in different ways.
They also remain compatible with a destructive or out-of-distribution response
to hard deletion. The comparison does not select among these explanations.

All of these statements are conditional on the fixed map. This is not a
same-dose contrast, and it does not calibrate the legacy outcomes onto a common
realized-dose scale or establish transport between the operators or
populations.
  
  \begin{figure*}[t]
  \centering
  \includegraphics[width=0.96\linewidth]{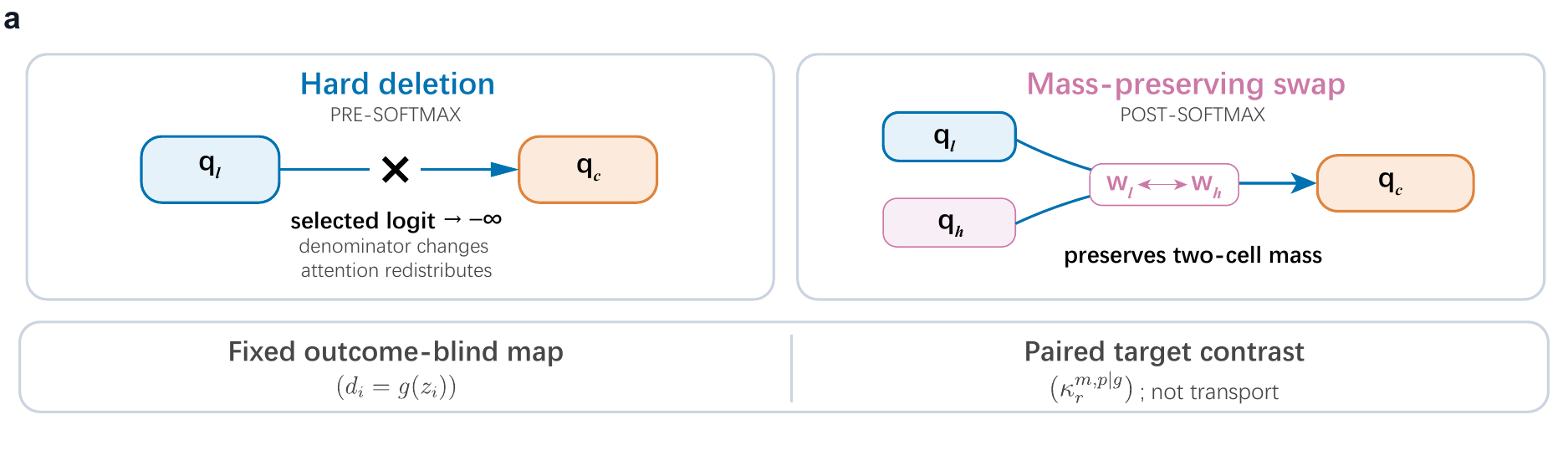}\\[0.6em]
  \includegraphics[width=0.96\linewidth]{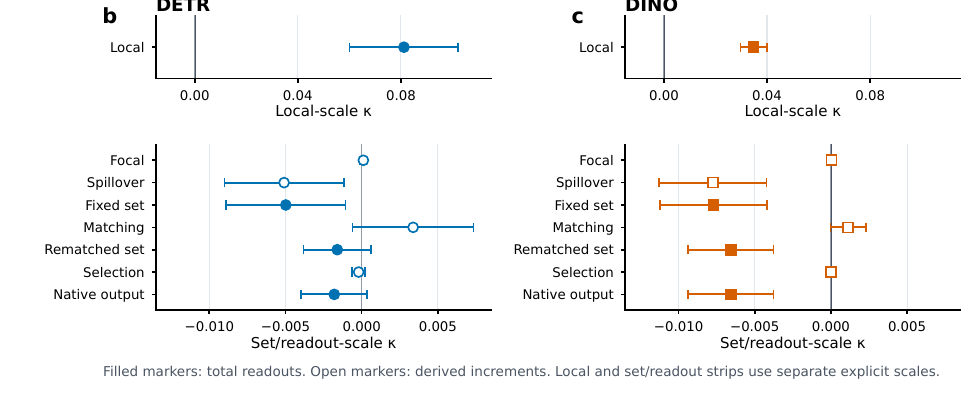}
  \caption{\textbf{Conditional mapped operator response.} Panel (a) contrasts
pre-softmax hard deletion with the post-softmax two-cell mass-preserving edit
under the fixed outcome-blind map. Panels (b,c) show the conditional mapped
contrast across local, derived, fixed-assignment, rematched, selection, and
native readouts. The comparison is not same-dose and does not identify operator
transport or a unique mechanism.}
\label{fig:operator}
\end{figure*}

\subsection{Dose and fresh-context evidence do not support a transport upgrade}

The completed legacy analysis contains all 1,420 expected model-image records,
and its integrity checks pass. The full recorded conjunction is not met at any
intermediate nominal dose. It combines target-arm local and set endpoints with
a target-versus-control joint-reversal component, so it is distinct from the
primary local-positive/fixed-assignment-negative estimand family.

Nominal attenuation is not a calibrated realized dose because the legacy
outcome records do not co-record per-head realized mass or immediate message
displacement. The legacy result therefore shows only that the tested conditions
do not pass their own conjunction; it does not establish formal non-transport.

The fresh assay records realized dose on 128 paired units per checkpoint. Its
pairwise intervals fall within the author-specified equivalence bands, but the
hard-sensitivity prerequisite fails in both checkpoints. Because the assay uses
a different population and primary-ground-truth formation process, and its hard
check uses a baseline rather than the matched active control used for the
discovery primary result, it cannot establish equivalence of the discovery
effect or population-wide non-transport. Calibrated operator transport remains
unresolved, and the planned mechanism upgrade is therefore not supported.

\subsection{Diagnostic sensitivity does not authorize a regularizer study}

The fixed-checkpoint design analysis covers 256 images per checkpoint. No
prespecified candidate satisfies the complete selection rule. The recorded
continuation therefore ends at diagnosis: no regularizer was trained or
evaluated, and the analysis supplies no evidence that static or adaptive
regularizers generally succeed or fail.

\section{Discussion}
\label{sec:discussion}

\subsection{Local improvement can coexist with set-level redistribution}

The central finding is a mismatch between two legitimate views of the same
edited computation. The local readout asks what happens at the selected
recipient; the fixed-assignment readout asks what happens to the set while
preserving the original correspondence. Their opposite signs show that a local
improvement need not be an isolated gain. In a jointly decoded system, the
response can be redistributed through other slots and later states. The finding
is a structured disagreement among readouts; it does not assign a natural
harmful role to the deleted relation or identify a unique causal path.

\subsection{Matching is part of the measured response}

Rematching is often treated as evaluation machinery applied after the model has
produced outputs. Here it changes what portion of the intervention response
remains visible. DETR's later-readout intervals cross zero, while DINO's remain
negative. Reassignment is therefore a recovery stage in the observed set-level
response, not a neutral relabeling. The fixed-assignment finding remains; the
rematched readout answers a different question about whether the altered set can
be reassigned to recover utility.

\subsection{Intervention versions define different scientific questions}

The hard and mass-preserving operators intervene at different locations and
change different computational quantities. Their fixed-map contrast appears
before rematching, which argues against explaining the entire difference through
the final assignment step. Yet the result cannot determine whether the source is
renormalization, donor content, later interaction, destructive deletion, or
another compatible pathway. These unresolved alternatives mark the point at
which the intervention response would need additional identification evidence
before becoming a mechanism claim.

The transport question remains open in the tested settings. A nominal dose is
not a realized dose when the quantities needed to connect them were not recorded
alongside the outcomes. The fresh assay does not close that gap because its
prerequisite fails and its population and baseline differ. Transport is therefore
unresolved, not ruled out.

\subsection{Diagnosis and design are separate empirical steps}

A selected fixed-checkpoint sensitivity can motivate a design hypothesis, but it
does not evaluate that hypothesis. The prespecified design analysis selects no
stable continuation, and no training occurs. A regularizer would change the
learned model, optimization trajectory, and distribution of query interactions;
it would not merely reproduce a fixed-checkpoint deletion. Treating the
diagnostic result as evidence for regularizer performance would therefore change
both the intervention and the estimand.

For intervention studies in shared prediction systems, the practical
implication is to report the edited computational location, comparator, selected
population, realized-dose contract, and readout stage together. Local, fixed,
rematched, and native outcomes should remain separate when shared computation
and assignment can change the output. Any move from fixed-checkpoint diagnosis
to training-time design requires its own authorization and evaluation.

\section{Limitations}

The primary results are conditional on selected image--relation populations from
two related checkpoints. They do not estimate arbitrary-relation prevalence or
an architecture-family effect. The active control edits a different recipient
and is therefore composite. The set utilities are not detector average
precision, and the study does not establish detector-level degradation.

The legacy outcomes lack co-recorded per-head realized dose and immediate
message displacement. The mapped hard-minus-mass comparison depends on a fixed
outcome-blind map and does not propagate mapping uncertainty. The fresh assay
changes context and fails its hard-sensitivity prerequisite. The evidence does
not identify a unique mechanism, establish calibrated operator transport or
population-wide non-transport, or evaluate a learned regularizer. Materials
reserved for prospective validation remain sealed and unread. The recorded
$0.30$ and $0.05$ IoU admission thresholds and the $10^{-4}$ event deadzone are
design choices of the evaluated protocol; threshold-sensitivity analyses were
not performed, so the reported populations and sign-event rates remain
conditional on them. The study also does not decompose the response by class,
object size, crowding, attention head, or decoder layer, and it does not provide
an AP bridge or qualitative image case study. These omissions limit claims about
heterogeneity, detector-level relevance, and when the checkpoint-specific
recovery pattern occurs.

\section{Conclusion}

Across two related shared set-decoder checkpoints, deleting a selected query
relation relative to a matched active control improves the local target slot
while reducing fixed-assignment set utility. Frequent within-unit reversals show
that this disagreement is not only an aggregate effect, and rematched/native
readouts show checkpoint-specific persistence. A conditional mapped hard-
minus-mass difference appears before rematching, but the available dose and
fresh-context evidence do not establish operator transport. The resulting
scientific claim is precise: the selected relations exhibit deletion sensitivity
whose observed consequence depends on the readout and intervention operator.
The evidence does not support an intervention-invariant mechanism or a training-
time design claim.

\section{Data and Code Availability}

Source code and the analysis-ready reproduction package are publicly available
at \url{https://github.com/yqjyzzz/query-interaction-reproduction}. The
repository provides per-image rows, analysis configurations, population and pair
manifests, integrity records, expected aggregates, and CPU-only code needed to
reproduce the reported aggregate analyses. The same frozen package is included
as arXiv ancillary material. Image pixels and model weights are not
redistributed; they must be obtained from their original sources and are not
required for aggregate reproduction. The package therefore supports exact
reconstruction of the reported statistics but not end-to-end model-inference
replication. A separately scoped detection demonstration in the repository does
not generate or replace any result reported in this paper. Materials reserved
for prospective validation were not accessed and are not included.

\section{Broader Impact}

More precise interpretation of internal interventions can reduce overclaiming
from positive ablations. The main risk is the opposite misreading: fixed-
assignment utility changes could be reported as detector-level degradation.
Keeping the comparator, selected population, and readout stage attached to each
result reduces both errors.

\paragraph{Funding and competing interests.}
This research received no specific grant from any funding agency in the public,
commercial, or not-for-profit sectors. Wuhan United Imaging Surgical Co., Ltd.
had no role in the study design, data analysis, interpretation of results, or
decision to submit the manuscript. The authors declare no competing interests.

\bibliographystyle{tmlr}
\bibliography{references_tmlr_verified}

@inproceedings{carion2020detr,
  title={End-to-End Object Detection with Transformers},
  author={Carion, Nicolas and Massa, Francisco and Synnaeve, Gabriel and Usunier, Nicolas and Kirillov, Alexander and Zagoruyko, Sergey},
  booktitle={Computer Vision -- ECCV 2020},
  pages={213--229},
  publisher={Springer International Publishing},
  year={2020},
  doi={10.1007/978-3-030-58452-8_13},
  url={https://arxiv.org/abs/2005.12872}
}

@inproceedings{jain2019attention,
  title={Attention is not Explanation},
  author={Jain, Sarthak and Wallace, Byron C.},
  booktitle={Proceedings of the 2019 Conference of the North American Chapter of the Association for Computational Linguistics: Human Language Technologies},
  pages={3543--3556},
  publisher={Association for Computational Linguistics},
  year={2019},
  doi={10.18653/v1/N19-1357},
  url={https://aclanthology.org/N19-1357/}
}

@inproceedings{wiegreffe2019attention,
  title={Attention is not not Explanation},
  author={Wiegreffe, Sarah and Pinter, Yuval},
  booktitle={Proceedings of the 2019 Conference on Empirical Methods in Natural Language Processing and the 9th International Joint Conference on Natural Language Processing},
  pages={11--20},
  publisher={Association for Computational Linguistics},
  year={2019},
  doi={10.18653/v1/D19-1002},
  url={https://aclanthology.org/D19-1002/}
}

@article{vanderweele2013versions,
  title={Causal Inference Under Multiple Versions of Treatment},
  author={VanderWeele, Tyler J. and Hern{\'a}n, Miguel A.},
  journal={Journal of Causal Inference},
  volume={1},
  number={1},
  pages={1--20},
  year={2013},
  doi={10.1515/jci-2012-0002},
  url={https://pmc.ncbi.nlm.nih.gov/articles/PMC4219328/}
}

@inproceedings{massidda2023soft,
  title={Causal Abstraction with Soft Interventions},
  author={Massidda, Riccardo and Geiger, Atticus and Icard, Thomas and Bacciu, Davide},
  booktitle={Proceedings of the Second Conference on Causal Learning and Reasoning},
  series={Proceedings of Machine Learning Research},
  volume={213},
  pages={68--87},
  publisher={PMLR},
  year={2023},
  url={https://proceedings.mlr.press/v213/massidda23a.html}
}

@inproceedings{li2022dndetr,
  title={DN-DETR: Accelerate DETR Training by Introducing Query DeNoising},
  author={Li, Feng and Zhang, Hao and Liu, Shilong and Guo, Jian and Ni, Lionel M. and Zhang, Lei},
  booktitle={IEEE/CVF Conference on Computer Vision and Pattern Recognition},
  pages={13619--13627},
  year={2022},
  doi={10.1109/CVPR52688.2022.01325},
  url={https://openaccess.thecvf.com/content/CVPR2022/html/Li_DN-DETR_Accelerate_DETR_Training_by_Introducing_Query_DeNoising_CVPR_2022_paper.html}
}

@inproceedings{zhang2023dino,
  title={DINO: DETR with Improved DeNoising Anchor Boxes for End-to-End Object Detection},
  author={Zhang, Hao and Li, Feng and Liu, Shilong and Zhang, Lei and Su, Hang and Zhu, Jun and Ni, Lionel M. and Shum, Heung-Yeung},
  booktitle={International Conference on Learning Representations},
  year={2023},
  url={https://openreview.net/forum?id=3mRwyG5one}
}

@inproceedings{chen2023groupdetr,
  title={Group DETR: Fast DETR Training with Group-Wise One-to-Many Assignment},
  author={Chen, Qiang and Chen, Xiaokang and Wang, Jian and Zhang, Shan and Yao, Kun and Feng, Haocheng and Han, Junyu and Ding, Errui and Zeng, Gang and Wang, Jingdong},
  booktitle={Proceedings of the IEEE/CVF International Conference on Computer Vision},
  pages={6633--6642},
  year={2023},
  doi={10.1109/ICCV51070.2023.00610},
  url={https://openaccess.thecvf.com/content/ICCV2023/html/Chen_Group_DETR_Fast_DETR_Training_with_Group-Wise_One-to-Many_Assignment_ICCV_2023_paper.html}
}

@inproceedings{gao2024ease,
  title={EASE-DETR: Easing the Competition among Object Queries},
  author={Gao, Yulu and Sun, Yifan and Ding, Xudong and Zhao, Chuyang and Liu, Si},
  booktitle={IEEE/CVF Conference on Computer Vision and Pattern Recognition},
  pages={17282--17291},
  year={2024},
  doi={10.1109/CVPR52733.2024.01636},
  url={https://openaccess.thecvf.com/content/CVPR2024/html/Gao_EASE-DETR_Easing_the_Competition_among_Object_Queries_CVPR_2024_paper.html}
}

@inproceedings{senthivel2024qrdetr,
  title={QR-DETR: Query Routing for Detection Transformer},
  author={Senthivel, Tharsan and Vu, Ngoc-Son},
  booktitle={Asian Conference on Computer Vision},
  pages={354--371},
  year={2024},
  doi={10.1007/978-981-96-0960-4_24},
  url={https://openaccess.thecvf.com/content/ACCV2024/html/Senthivel_QR-DETR__Query_Routing_for_Detection_Transformer_ACCV_2024_paper.html}
}

@inproceedings{liu2023stable,
  title={Detection Transformer with Stable Matching},
  author={Liu, Shilong and Ren, Tianhe and Chen, Jiayu and Zeng, Zhaoyang and Zhang, Hao and Li, Feng and Li, Hongyang and Huang, Jun and Su, Hang and Zhu, Jun and Zhang, Lei},
  booktitle={Proceedings of the IEEE/CVF International Conference on Computer Vision},
  pages={6491--6500},
  year={2023},
  doi={10.1109/ICCV51070.2023.00597},
  url={https://openaccess.thecvf.com/content/ICCV2023/html/Liu_Detection_Transformer_with_Stable_Matching_ICCV_2023_paper.html}
}

@inproceedings{hou2024relation,
  title={Relation DETR: Exploring Explicit Position Relation Prior for Object Detection},
  author={Hou, Xiuquan and Liu, Meiqin and Zhang, Senlin and Wei, Ping and Chen, Badong and Lan, Xuguang},
  booktitle={Computer Vision -- ECCV 2024},
  pages={89--105},
  year={2024},
  publisher={Springer Nature Switzerland},
  doi={10.1007/978-3-031-72973-7_6},
  url={https://www.ecva.net/papers/eccv_2024/papers_ECCV/papers/06646.pdf}
}

@misc{zhang2026dualr,
  title={Dual-R-DETR: Resolving Query Competition with Pairwise Routing in Transformer Decoders},
  author={Zhang, Ye and Chen, Qi and Huang, Wenyou and Liu, Rui and Kang, Zhengjian},
  year={2026},
  note={arXiv preprint arXiv:2512.13876; accepted at ICME 2026},
  doi={10.48550/arXiv.2512.13876},
  url={https://arxiv.org/abs/2512.13876}
}

@inproceedings{zhang2024patching,
  title={Towards Best Practices of Activation Patching in Language Models: Metrics and Methods},
  author={Zhang, Fred and Nanda, Neel},
  booktitle={International Conference on Learning Representations},
  year={2024},
  url={https://proceedings.iclr.cc/paper_files/paper/2024/hash/06a52a54c8ee03cd86771136bc91eb1f-Abstract-Conference.html}
}

@inproceedings{makelov2024subspace,
  title={Is This the Subspace You Are Looking for? An Interpretability Illusion for Subspace Activation Patching},
  author={Makelov, Aleksandar and Lange, Georg and Geiger, Atticus and Nanda, Neel},
  booktitle={International Conference on Learning Representations},
  year={2024},
  url={https://proceedings.iclr.cc/paper_files/paper/2024/hash/70b8505ac79e3e131756f793cd80eb8d-Abstract-Conference.html}
}

@misc{quirke2026transfer,
  title={Ablation-Reversible Heads Don't Transfer: A Stress Test for Mechanistic Role Claims in Transformers},
  author={Quirke, Philip},
  year={2026},
  note={arXiv preprint arXiv:2606.08292},
  doi={10.48550/arXiv.2606.08292},
  url={https://arxiv.org/abs/2606.08292}
}

@misc{joshi2026generalise,
  title={Causality is Key for Interpretability Claims to Generalise},
  author={Joshi, Shruti and Mueller, Aaron and Klindt, David and Brendel, Wieland and Reizinger, Patrik and Sridhar, Dhanya},
  year={2026},
  note={arXiv preprint arXiv:2602.16698},
  doi={10.48550/arXiv.2602.16698},
  url={https://arxiv.org/abs/2602.16698}
}

@misc{guo2026igsd,
  title={Beyond Importance: Interchange-Sobol Sensitivity Reveals Task-Specific Content Channels in Transformer Components},
  author={Guo, Yifeng and Du, Jin-Hong and Chen, Xiang},
  year={2026},
  note={arXiv preprint arXiv:2606.20678},
  doi={10.48550/arXiv.2606.20678},
  url={https://arxiv.org/abs/2606.20678}
}

\appendix
\section{Supporting Analyses and Provenance}
\label{app:supporting}

\subsection{Operator and context checks}

The completed legacy dose analysis contains 1,420 intervention records. Its
identity, repeated-baseline, and zero-dose controls are exactly zero, but no
intermediate mass-preserving condition passes the prespecified conjunction of
local, set-level, and joint-event criteria. The original records do not contain
per-head realized dose or immediate message displacement, so this result does
not establish calibrated operator transport.

An input-only replay on the discovery inputs records target and donor weights
and immediate message displacement without reading scientific outcomes. Its
overlap check supports the fixed map used for the conditional hard-minus-mass
analysis. The separate COCO assay records realized dose on 128 paired units per
checkpoint. Its pairwise operator intervals fall inside the assay-specific
bands, but the required hard-deletion sensitivity check fails in both
checkpoints. These analyses leave the discovery effect's operator transport
unresolved.

A final fixed-checkpoint analysis evaluates whether the diagnostic evidence is
sufficient to justify a proposed static regularizer study. None of its complete
prespecified candidates passes on 256 images per checkpoint, so no training study is
undertaken. This is a design-stopping result, not evidence about regularizer
performance.

\subsection{Native endpoint provenance note}

In this note, H4-D and T0 are historical provenance aliases: H4-D denotes the
decomposition-specific analysis record, whereas T0 denotes the common-estimand
audit. Neither alias is a canonical endpoint.

The native readout reported in H4-D is a derived decomposition-specific summary
and is not identical to the native endpoint defined in the T0 common estimand
audit. Because these endpoints differ in estimand construction and analytical
purpose, we do not combine, compare, or substitute their numerical values. All
primary quantitative claims use the T0-defined estimands.

\section{Operator and Population Details}
\label{app:implementation}

The hard-deletion hook is the decoder input layer-3 pre-softmax attention-logit
tensor. For each intervened head, the selected target logit is set to
$-\infty$ before softmax. The denominator therefore changes and deleted mass can
be redistributed over remaining sources. The mass-preserving arm edits the
corresponding post-softmax target/donor weights at the same layer while
preserving their two-cell mass.

\begin{table}[h]
\centering
\small
\caption{\textbf{Intervention-arm mapping at decoder layer 3.} Arrows name
recipients, sources, and donors; target and active-control arms use different
recipients.}
\label{tab:mapping}
\begin{tabular}{lll}
\toprule
Arm & Recipient & Source/donor rule \\
\midrule
Hard target & competitor & delete leader source before softmax \\
Hard active control & sham & delete same leader source at different recipient \\
Mass target & competitor & swap leader source with sham donor \\
Mass placebo & sham & swap leader source with competitor donor \\
\bottomrule
\end{tabular}
\end{table}

The discovery and supporting assays use the same target-pair rule but different population,
dataset, primary-ground-truth formation, ranking, and admission contexts. The
denominator is therefore selection-conditional.
The intervention-arm mapping is summarized in Table~\ref{tab:mapping}.

\section{Readout and Reproducibility Contract}
\label{app:provenance}

The fixed-assignment, rematched, and native readouts are distinct endpoints.
The decomposition-specific native summary and the common-estimand native
endpoint are not combined or substituted. The supplementary manifest records
the complete hashes, receipt schemas, statistical configurations, and
implementation files used for aggregate reconstruction.

\begin{figure}[h]
\centering
\includegraphics[width=0.84\linewidth]{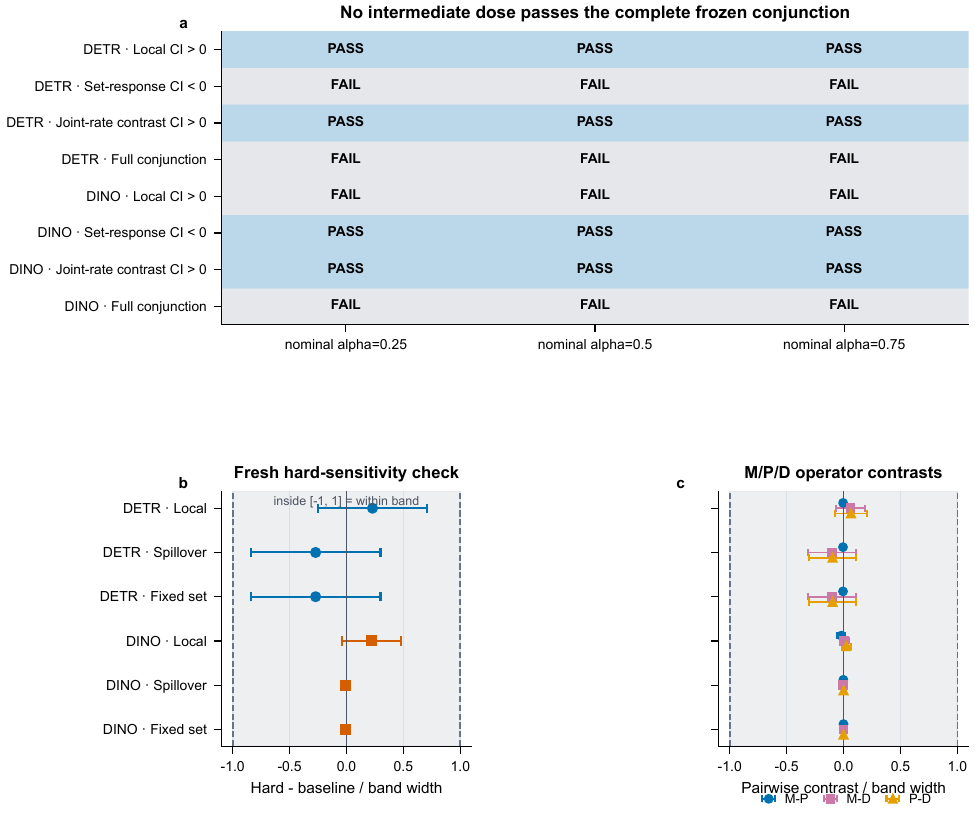}
\caption{\textbf{Supporting operator and context checks.} The legacy
mass-preserving conditions fail different components of the declared
conjunction, while the fresh assay fails its required hard-deletion sensitivity
check. These results do not establish population non-transport or regularizer
performance.}
\label{fig:transport}
\end{figure}

Figure~\ref{fig:transport} summarizes the legacy dose conjunction and the fresh
hard-sensitivity check.

\end{document}